\documentclass[jair,twoside,11pt,theapa]{article}
\usepackage{jair, theapa, rawfonts, xcolor, graphicx}
\usepackage{enumerate,enumitem}

\begin{document}
\jairheading{1}{2024}{1-15}{2024}{Y}

\title{The future of human-centric eXplainable Artificial Intelligence (XAI) is not post-hoc explanations}

\author{\name Vinitra Swamy \email vinitra.swamy@epfl.ch \\
       \name Jibril Frej \email jibril.frej@epfl.ch \\
       \name Tanja Käser \email tanja.kaeser@epfl.ch \\
       \addr EPFL, Switzerland}

\maketitle

\begin{abstract}
Explainable Artificial Intelligence (XAI) plays a crucial role in enabling human understanding and trust in deep learning systems. As models get larger, more ubiquitous, and pervasive in aspects of daily life, explainability is necessary to minimize adverse effects of model mistakes. Unfortunately, current approaches in human-centric XAI (e.g. predictive tasks in healthcare, education, or personalized ads) tend to rely on a single post-hoc explainer, whereas recent work has identified systematic disagreement between post-hoc explainers when applied to the same instances of underlying black-box models. In this paper, we therefore present a call for action to address the limitations of current state-of-the-art explainers. We propose a shift from post-hoc explainability to designing interpretable neural network architectures. We identify five needs of human-centric XAI (real-time, accurate, actionable, human-interpretable, and consistent) and propose two schemes for interpretable-by-design neural network workflows (adaptive routing with InterpretCC and temporal diagnostics with I2MD). We postulate that the future of human-centric XAI is neither in explaining black-boxes nor in reverting to traditional, interpretable models, but in neural networks that are intrinsically interpretable.

\end{abstract}

\section{Introduction}
The rise of neural networks is accompanied by a severe disadvantage: the lack of transparency of their decisions. Deep models are often considered black-boxes, producing highly accurate results while providing little insight into how they arrive at those conclusions. This disadvantage is especially relevant in human-centric domains where model decisions have large, real-world impacts \shortcite{webb2021machine,conati2018ai}.

\begin{figure*}[t]
  \centering
  \includegraphics[width=\textwidth]{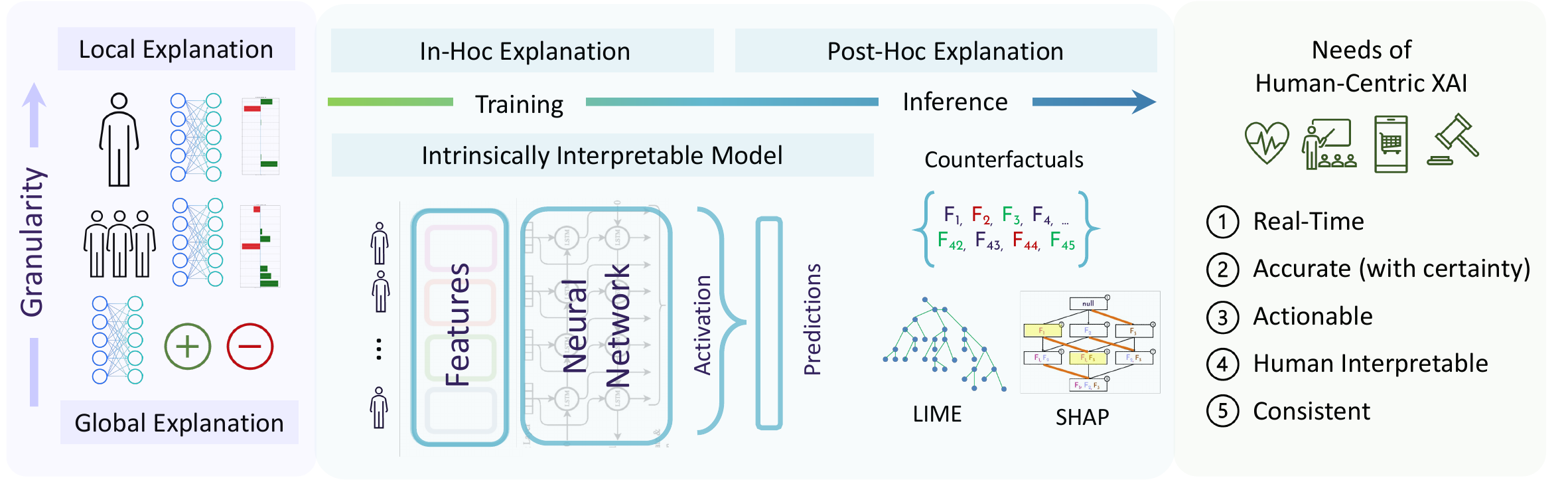}
  \caption{Explainability can be intrinsic (by design), in-hoc (e.g., gradient methods), or post-hoc (e.g., LIME, SHAP). Furthermore, the granularity of model explanations ranges from local (single user, a group of users) to global (entire sample).}
  \vspace{-8mm}
  \label{fig:pipeline}
\end{figure*}

The goal of eXplainable AI (XAI) is to circumvent this failing by either producing interpretations for black-box model decisions or making the model's decision-making process transparent. As illustrated in Figure \ref{fig:pipeline}, model explanations range from local (single point) to global granularity (entire sample). Moreover, explainability can be integrated into the modeling pipeline at three stages: 
\begin{enumerate}[leftmargin=*]
    \item \textbf{Intrinsic explainability}: traditional ML models (e.g., decision trees) explicitly define the decision pathway.
    \item \textbf{In-hoc explainability}: interpreting the model gradients at inference or customizing training protocols for additional information; e.g., Grad-CAM uses backpropagation to highlight important regions of an input image \shortcite{selvaraju2017grad}.
    \item \textbf{Post-hoc explainability}: after the decision is made, an explainer is fit on top of the black-box model to interpret the results.
\end{enumerate}
  
In human-centric domains, researchers and practitioners tend to use either intrinsically interpretable traditional ML models \shortcite{jovanovic2016building,vultureanu2021improving} or apply a single post-hoc explainer \shortcite{adadi2018peeking,dovsilovic2018explainable}. Unfortunately, recent research shows that post-hoc explanations might be unfaithful to the true model \shortcite{rudin2019stop}, inconsistent \shortcite{slack2020fooling}, or method-dependent \shortcite{Swamyexplainers2022,krishna2022disagreement,brughmans2023disagreement}. Furthermore, evaluating the quality of the provided explanations is a challenge, since there is often no ground truth \shortcite{swamy2023trusting,dai2022fairness}. 

In this paper, we therefore present a call-to-action to address the limitations of current state-of-the-art explainability methods. While previous work \shortcite{rudin2019stop} has made a strong argument for moving away from black-box models and using inherent interpretability (i.e. traditional ML models) for impactful decisions, we suggest exploring strategies to make \textit{deep learning} approaches intrinsically interpretable, guaranteeing transparency, robustness, and trustworthiness. We believe that human-centric domains should profit from both explainability and the recent advances in state-of-the-art machine learning methods, including large language models (LLMs).

In the following, we define five needs of human-centric XAI: real-time, accurate, actionable, human-interpretable, and consistent. We discuss the limitations of current XAI methods, their inability to meet the requirements for human-centric XAI, and two ideas towards inherently interpretable deep learning workflows. 
We hope this paper will serve as a guideline for achieving consistency and reliability in human-centric XAI systems.

\section{Requirements for Human-Centric eXplainable AI}
Neural networks have an enormous potential for impacting human life, from areas like personalized healthcare or educational tutoring to smart farming and finance. 
We define human-centric as any application where a human directly uses model predictions in decision-making.
In light of the specific challenges in human-centric domains \shortcite{national2021human}, we define five requirements that explanations should fulfill.
\begin{enumerate}[leftmargin=*]
    \small
    \item\textbf{Real-Time}: Explanations should be provided in real-time or with minimal delay to support timely decision-making (in the scale of seconds, not tens of minutes), e.g., \shortciteA{xu2017real}.
    \item \textbf{Accurate explanations with certainty}: Explanations need to be accurate, reflecting the neural network's decision-making process or at least accompanied by a level of confidence \shortcite{marx2023but,leichtmann2023effects}.
    \item \textbf{Actionable}: Explanations should provide actionable insights, empowering model deployers to take appropriate actions or make informed interventions \shortcite{joshi2019towards}.
    \item \textbf{Human interpretable}: Explanations should be understandable to a broad audience beyond computer scientists \shortcite{hudon2021explainable,haque2023explainable}. We believe that LLMs will be crucial in improving the understandability of explanations. 
    \item \textbf{Consistent}: Explanations should be consistent across similar instances or contexts, ensuring reliability and predictability in the decision-making process. In a time series of interactive predictions, the explanations should not drastically differ \shortcite{li2021algorithmic}.
\end{enumerate}

\section{Explainers of Today: State-of-the-Art and Limitations}

Research and adoption of neural network explainability in human-centric areas has surged over the last eight years. In-hoc methods like layer relevance propagation \shortcite{lu2020towards} or concept-activation vectors \shortcite{cav} have shown success in student success prediction \shortcite{asadi2022ripple} or identifying skin conditions \shortcite{lucieri2020interpretability}, but require specific model architectures or access to model weights. Intrinsic explainers like neural additive models \shortcite{agarwal2021neural} have shown aptitude for personalized treatments of COVID-19 patients, but require developer effort and could affect model performance. Post-hoc approaches are most commonly favored, as there is no impact on model accuracy and no additional effort required during training. Local, instance-specific post-hoc techniques such as LIME \shortcite{lime}, SHAP \shortcite{shap}, or counterfactuals \shortcite{dice}, have been effectively utilized for tasks like predicting ICU mortality \shortcite{katuwal2016machine}, non-invasive ventilation for ALS patients \shortcite{ferreira2021predictive}, credit risk \shortcite{creditrisk}, or loan repayment \shortcite{pawelczyk2020learning}.

Post-hoc approaches, while popular, are accompanied by weaknesses in real-world settings. The computational time is often in the tens of minutes; not \textbf{real-time} enough for users, students, or patients to make a decision based on the explanation alongside a prediction. In most cases, there is \textbf{no measurement of confidence} in a generated post-hoc explanation. The \textbf{actionability} and \textbf{human-interpretability} of the explanation are solely based on the input format. As human-centric tasks often use tabular or time series data, the subsequent explanations are often not concise, actionable or interpretable easily beyond the scope of a data scientist's knowledge \cite{karran2022designing}. Recent research on explanation user design has shown that humans across healthcare, law, finance, education, and e-commerce, among others, prefer hybrid text and visual explanations \shortcite{haque2023explainable}, a format not easily provided by current post-hoc libraries. Lastly, the \textbf{consistency} of the explanations is not inherently measured. Several explainability methods could produce vastly different explanations with different random seeds or at different time steps \shortcite{slack2020fooling}.

Furthermore, post-hoc explanations are difficult to evaluate. Current metrics (e.g. saliency, faithfulness) aim to quantify the quality of an explanation in comparison to expert-generated ground truth \cite{agarwal2022openxai}. However, accurate explanations need to be true to the model internals, not human perceptions. In this light, the most trustworthy metrics measure the prediction gap (e.g. PIU, PGU), removing features that are considered important by the explanation and seeing how the prediction changes \shortcite{dai2022fairness}. This approach is still time-consuming and imperfect, as it fails to account for cross-feature dependencies. Recent literature \shortcite{krishna2022disagreement,brughmans2023disagreement,Swamyexplainers2022} has examined the results of over $50$ explainability methods with diverse datasets ranging from criminal justice to healthcare to education through a variety of metrics (rank agreement, Jenson-Shannon distance) and demonstrated strong, systematic disagreement across methods. Validating explanations through human experts can also be difficult: explanations are subjective, and most can be justified. \shortciteA{krishna2022disagreement}, \shortciteA{swamy2023trusting}, and \shortciteA{dhurandhar2018explanations} have conducted user studies to examine trust in explainers, measuring data scientist and human expert preference of explanations. Results indicate that while humans generally find explanations helpful, no method is recognized as most trustworthy. As further shown by \shortciteA{swamy2023trusting}, most preferred explanations align with the prior beliefs of validators. 

We anticipate that the state-of-the-art in AI will continue to prefer large, pretrained deep models over traditional interpretable models for the foreseeable future; the capabilities and ease-of-use of neural networks outweigh any black-box drawbacks. Our goal is therefore to identify a way to use deep learning in an interpretable workflow.

\section{Intrinsically Interpretable Deep Learning Design}
In human-centric applications, there is no margin for error; It is crucial to prioritize designs that are intrinsically interpretable as opposed to imperfect approximations of importance. We present two ideas towards intrinsically interpretable deep learning workflows (InterpretCC and I2MD), targeting both local and global explanability.

\begin{figure*}[h]
  \centering
  \includegraphics[width=\textwidth]{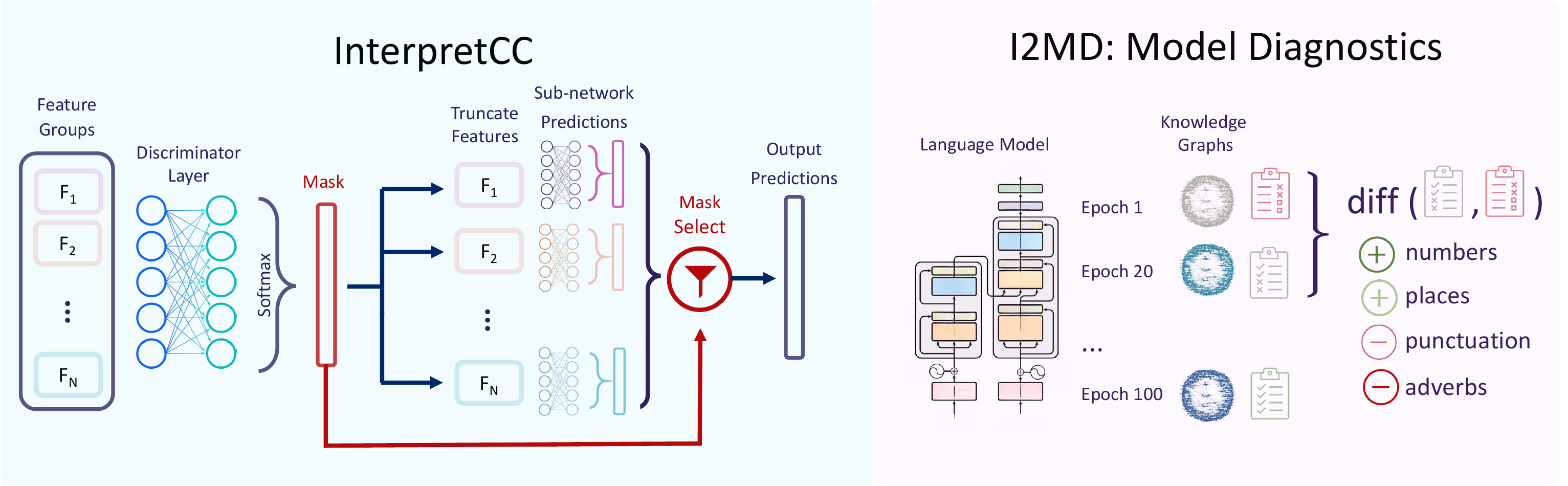}
  \caption{Proposed architecture of adaptive routing with Intepretable Conditional Computation (InterpretCC, left). A discriminator layer adaptively selects feature groupings as important, then sends truncated feature sets to expert sub-networks. Example of global model benchmarks with Interpretable Iterative Model Diagnostics (I2MD, right). Knowledge graphs are extracted from a language model at iterative stages of training and compared over time with diagnostic benchmarks.}
  \vspace{-8mm}
  \label{fig:ideas}
\end{figure*}

\subsection{InterpretCC: Interpretable Conditional Computation}
InterpretCC aims to guarantee an explanation's accuracy to model behavior with $100\%$ certainty, while maintaining performance by input point adaptivity. This approach is inspired by conditional computation in neural networks \shortcite{bengio2013estimating} to speed up neural network computation. While it is similar to feature grouping interpretability approaches like \shortciteA{nauta2023pip,chen2018looks}, it differentiates by its focus on human-specified feature groupings and an adaptive choice of expert subnetworks. Extensive experiments of InterpretCC have been performed in \shortciteA{swamy2024interpretcc}.

The simplest implementation of InterpretCC is a feature gating model that learns a dynamic feature mask and enforces sparsity regularization on the input features. For each point, the goal is to choose a minimal predictive feature set. While it might seem that model accuracy will be compromised by this approach, the adaptivity has potential to improve performance by reducing noise. This idea can be expanded to an interpretable mixture-of-experts model by dynamically activating expert sub-networks (Figure \ref{fig:ideas}). Instead of restricting the features individually, we can group features together meaningfully (either by human specification or automated approaches). Expert sub-networks are trained only using their subset of features and either activated or ignored for each point. 

InterpretCC optimizes the interpretability-accuracy trade-off: easy-to-classify instances use less features and therefore have high interpretability while difficult-to-classify points use more features and do not trade accuracy for interpretability. The advantages of InterpretCC are multifold, as explanations are 1) \textbf{real-time} (a prediction is provided simultaneously with the explanation) 2) \textbf{accurate} (the model only uses specific features or feature groups), 3) \textbf{consistent} (the same learned experts will be activated for each point), and 4) \textbf{human interpretable} (sparse explanations with human-specified groupings of features). InterpretCC's \textbf{actionability} depends on the actionability of the user-specified features (e.g., in the breast cancer setting of \shortciteA{swamy2024interpretcc} with grouped cell nuclei, explanations are not actionable).

\subsection{I2MD: Interpretable Iterative Model Diagnostics}

Current deep learning performance metrics (accuracy, F1 score) paint a starkly incomplete global picture of model strengths and weaknesses. I2MD seeks to address this gap by examining the differential diagnostics of iterative model snapshots during training to build a detailed understanding of model abilities. Experimental evaluation of this approach in \shortciteA{swamy2021interpreting,liu2021probing} demonstrates
how language models can be interpreted by comparing knowledge graphs extracted during various stages of training (Figure \ref{fig:ideas}). This enables understanding of which skills the model learns at what time. During training or fine-tuning, tailored I2MD datasets can be created to target extracted model weaknesses; this results in a more performant model earlier in the training process and closes the loop, integrating XAI results back into the modeling pipeline. While the idea of iterative temporal diagnostics has been extensively discussed for usability \shortcite{hewett1986role}, its role in interpretability has not been explored yet.

The I2MD approach provides explanations that are \textbf{consistent} (a model snapshot will extract the same diagnostic explanations every time) and \textbf{actionable} (granular benchmarking allows developers to correct their models with custom datasets). However, it is not \textbf{real-time}, as extracting diagnostics from model snapshots is time-consuming in the training process. I2MD's \textbf{human interpretability} depends on the choice and granularity of diagnostics. Likewise, \textbf{accuracy} depends on the breadth of the diagnostics chosen and does not have a measure of certainty. A narrow iterative benchmark might not fully capture model weaknesses, while an overly broad iterative benchmark might not be easily understandable, illustrating the interpretability-accuracy tradeoff.

\section{Conclusion}
The evolving landscape of machine learning models, characterized by the ubiquity of LLMs, transformers, and other advanced techniques, necessitates a departure from the traditional approach of explaining black-box models. Instead, there is a growing need to incorporate interpretability as an inherent feature of model design. In this work, we have discussed five needs of human-centric XAI and have shown that the current state-of-the-art is not meeting these needs. We have also presented two initial ideas towards intrinsic interpretable design for neural networks and discussed their applications towards the five needs of human-centric XAI. As researchers, model developers, and practitioners, we must move away from imperfect, post-hoc XAI estimation and towards guaranteed interpretability with less friction and higher adoption in deep learning workflows.

\vskip 0.2in
\bibliography{citations}
\bibliographystyle{theapa}

\end{document}